*Review*

# Artificial Intelligence in Intelligent Tutoring Robots: A Systematic Review and Design Guidelines

**Jinyu Yang** [1], **Bo Zhang** [2]

1. School of Computers, National University of Defense Technology; dr.jinyu.yang@outlook.com
2. Artificial Intelligence Research Center, National Innovation Institute of Defense Technology; bo.zhang.airc@outlook.com
* Correspondence: bo.zhang.airc@outlook.com

**Abstract:** This study provides a systematic review of the recent advances in designing the intelligent tutoring robot (ITR), and summarises the status quo of applying artificial intelligence (AI) techniques. We first analyse the environment of the ITR and propose a relationship model for describing interactions of ITR with the students, the social milieu and the curriculum. Then, we transform the relationship model into the perception-planning-action model for exploring what AI techniques are suitable to be applied in the ITR. This article provides insights on promoting human-robot teaching-learning process and AI-assisted educational techniques, illustrating the design guidelines and future research perspectives in intelligent tutoring robots.

## 1. Introduction

The recent advances in Artificial Intelligence (AI) techniques have attracted great contributions from the academia and industry. Powered by the dramatically increased computational power and available data, the deep neural network based machine learning techniques are prospering and being applied to facilitate more intelligent robots [1]. Robots have been applied widely in our daily lives, and the number of service robots has already surpassed that of industrial robots in 2008 [2]. Robots are slowly beginning to integrate in the daily life. Social robots have played an even more important role in children and young people' lives as robots can be applied to promote their development and intellect.

Education is primarily related to understanding and supporting teaching and learning. It focuses on how to teach and learn as both of them are impacted by communication, course and curriculum design, assessment, and motivation. The continuous advent of new technology and rapid advancement of AI techniques can contribute to improve and enrich educational methods. AI techniques may find ways to enhance the acquisition, manipulation, and utilization of knowledge and the conditions where learning takes place, hence may help educators improve their effective teaching and promote students' individualised learning. Therefore, it is essential to understand what and how AI techniques can be used to achieve educational goals, namely, producing accessible, affordable, efficient, and effective teaching is a long-term goal of education [3].

The aim of this article is to provide an overview and build a bridge between education and robots with AI techniques as there is limited work on a comprehensive overview of the education robots with AI techniques in the literature. Woolf has worked on a book to provide a comprehensive overview on intelligent tutors in the field of robots [3], however, the robots and techniques in their book are not the state of the art as the AI techniques have been developing rapidly in these ten years. In recent years, many researchers have worked on the specific parts of the intelligent tutor. The authors of [4] provide an overview of empirical evidence for understanding current knowledge on learning analytics and educational data mining and its impact on adaptive learning. The authors of [5] review the adaptive system in education, and compare intelligent tutoring systems with adaptive hypermedia systems, and show examples of the implementation of the two kinds of system. Truong has reviewed articles in integrating learning styles and adaptive e-learning system [6]. In this article, we not only provide a whole picture of the intelligent tutoring robots with the state of the art AI techniques, but also build a framework for understanding and designing the intelligent tutoring robot (ITR) with AI techniques. Specifically, we use the perception-planning-action framework to fill up the gap between the teaching-learning relationship analysis in education domain and the ITR design in AI and robotic domain.

Human teachers have achieved excellent results in teaching, so it is useful to start to explore intelligent tutors through observations on human teachers. This paper firstly discusses teacher competences in teaching activities, and describes the relations of the four factors involved: teacher, student, curriculum and social milieu. Then, it proposes a framework of designing intelligent tutor systems using artificial intelligence techniques, along with reviews and design guidelines of the recent advances that can be used in the systems. We transform the relationship model of teachers' role in teaching activities into the well-established perception-planning-action model in the area of AI. The final section provides insights on future research on ITRs with AI techniques.

## 2. Relationship Model of the Teaching-Learning Process

This section presents three dimensions of teacher competences in teaching activities, which we can use AI techniques to represent in intelligent tutoring robots. How the ITR can replace a human teacher? How can these AI techniques improve teaching? In order to find the answers, we investigate ITRs from the perspective of human teachers.

There are heat debates on "what teachers should know" from a century ago as a theoretical and empirical discussion. In 1897, John Dewey suggested that educational process is complicated with both psychological and sociological sides, and educators are required to understand and have knowledge of multiple domains of teaching and learning [7].

Drawing on Schwab's "commonplaces of educating", the analysis of the elements of teacher competence has formed interaction groups [9]. Schwab believes that teaching activities are composed of four commonplaces, namely, teachers, students and curriculum that are all influenced by the social environment where educating takes place. These elements interact and develop into five commonplace groups: The teacher-self, the teacher-social milieu, the teacher-curriculum, the student-curriculum, and the teacher-student [9]. Also, the author of [10] places the teacher competence into classroom context and puts forward ten factors. These factors include pedagogy, content, curriculum, pedagogy content knowledge, interpersonal, intrapersonal, knowledge of students, adaptive expertise, and social responsibility [8,9,11-14]. While teacher capacity shows the knowledge, skills, and dispositions which a successful teacher must have, the commonplaces of educating demonstrates the experience of teaching, and the events that occur during educating [9]. These interaction groups are combined with the ten factors of teacher capacity to form a framework of teachers' role in teaching activities [10]. Our paper focuses on the intelligent tutor with AI techniques, so we only list the dimensions and skills that can be applied and developed in the intelligent tutor. According to [8,12,13], human teachers not only interact with the student, the curriculum, and the social milieu, but also with themselves. Therefore, teacher capacity includes the ability for introspection and reflection, which produces confidence and a sense of self-image for teachers [8,12,13]. However, most robots or intelligent tutors did not have such a function, and it is difficult for robots to reflect themselves. In this case, ITRs need to seek guidance from the ITR designers and human tutors. Also, there is some overlap between some interaction groups, so we combine the interaction group, students and curriculum, with other interaction groups. Therefore, this article discusses the three dimensions of the teaching activities: **teacher and student, teacher and social milieu, and teacher and curriculum.**

*3.1. Teacher and student*

The first two important elements in teaching activities are the teacher and the learner. What capacity is needed in this interaction group?

- Pedagogy: First of all, pedagogy is an essential part of a teacher's professional knowledge during educating process [11-13], which is a general body of knowledge about learning, instructing, and learners[10]. Pedagogy includes practical aspects of teaching, curricular issues, and the theoretical fundamentals of how and why learning occurs [11].
- Students Diversity Awareness: Teachers need to understand students and take into account the characteristics of the students, such as race, religion, physical characteristics, personal life choices (clothing, food, music, lifestyle), cultural factors (clothing, food, music, rituals) and body image, or cognitive diversity, such as personality or learning differences [8,11,13]. Also, there are emotional, mental, or physical diversities in students, so it is a teacher's responsibility to understand these differences and promote tolerance, curiosity, and equity among their students [8].
- Responding to students: Teachers need to be aware of students' emotion and response during teaching, so they can respond to them accordingly. Experienced teachers can easily distinguish between active students (taking notes or preparing to make comments) and passive students (too tired or bored to participate) with a quick glance [3]. Teachers are also required to respond the current culture and community in the teaching process and make efforts to promote learning with all students, in particular those with lower scores on the tests for accountability [11]. In addition, teachers should understand learners' thinking, the factors that make it easy or difficult for students to learn a particular topic and the best way to teach content to students [11]. Students of different ages and backgrounds bring their own concepts and preconceptions into the process of learning the most regularly taught topics and lessons [11]. If those preconceptions hurdle the learning of new knowledge, teachers need to master the strategies that can reorganise the learners' understanding [11].
- Multiple communication methods: Without communication, teaching cannot take place. Communication is essential as teachers use it to deliver lessons, convey concepts, understand students' knowledge and motivate students. Using communication strategies and a wide range of methods such as analysing written work, providing explanations and drawing graphics [3], human teachers can effectively develop knowledge of students and transmit information.
- Building relationships: A human teacher is supposed to communicate with students and others, and build relationships with them in order to create a community of learners [16].

*3.2. Teacher and social milieu*

This interaction group includes factors of social responsibility and adaptive expertise. A key feature of teacher capacity is that teachers need to be able to develop themselves, be flexible, and adapt to changing situations over time [8,12]. This is because the world and its associated knowledge are changing frequently and rapidly [8]. Teachers also need to realise that learning to teach is continuous [12,14]. "On-going" learning experience helps teachers grow and change as adaptive expertise and context-solving skills cannot be simulated [8]. Therefore, it is important for teachers to understand their social environment and their sense of responsibility as a professional in this environment. For instructor robots, this is the same situation.

*3.3. Teacher and curriculum*

Content knowledge, also as domain knowledge, is a necessary part of a teacher's professional knowledge [11,12]. The knowledge represents expert knowledge, or how experts perform in the domain, including definitions, processes, or skills needed [3]. Teachers are expected to understand the content or subject matter, the facts and concepts, substantive knowledge and the framework for organising these facts and concepts, and syntactic knowledge as well as the methods and means of obtaining subject knowledge [9,13,14]

Teachers will not only have the knowledge of a domain, but also grasp the curriculum designed for students, and national and local standards to evaluate students' learning performance [14]. Researchers define curriculum as the content and norms of a course or program of study [11]. An ideal teacher will "use a variety of open-ended, applied projects so children can practice the reflective process, link subject matter to real-life situations, and think of creative products to demonstrate their learning" [12](p.11) [13]. Teachers also should have the ability to critically evaluate the curricular, and look for or design teaching materials that suit to their students [11]. National, State and Local standards serve as a means of evaluating the performance or quality of work that students must achieve [14]. Competent teachers will have knowledge of these standards and be able to organise teaching within the curriculum to achieve the objectives of these standards [14].

Human teachers are also expected to grasp the ways of **expressing** and **developing** contents in their subjects that enable learners to comprehend them [11]. This means teachers need to **possess the most frequently taught topics**, the most useful expressions of those ideas, "the most powerful **analogies**, **illustrations**, **examples** and **demonstrations** in their subject area" [11](p.9-10). Teachers are also expected to have an understanding of "the curricular alternatives and materials for instruction, the alternative texts, software, programs, visual materials, single-concept films, laboratory demonstrations, or 'invitations to inquiry' [11](p.10).

There are many types of knowledge (topics, misconceptions and bugs, affective characteristics, student experience, and stereotypes), and teachers are expected to teach in a variety of ways [3]. It may take many years for human teachers to develop pedagogical content knowledge. However, the ITR may develop such expertise within much shorter time with the aid of state-of-the-art AI techniques powered by the dramatically increased data and computational resources.

## 3. Artificial Intelligence techniques for Designing Intelligent Tutor Robots

Following the relationship model analysed in the last section, this section devotes to the analysis of applying AI techniques in building and evaluating the ITR, along with a review of the state-of-art research contributions. In order to fill up the gap between the education domain and artificial intelligence domain, this section first transforms the relationship model describing the teaching-learning process, into the well-established perception-planning-action model in the area of AI. Then, we analyse the architecture of ITR, and then discuss how AI techniques may be applied to the system architecture, module design as well as system evaluation. It is worth noting that Woolf approaches the issue from a different perspective, in which ITRs encode student knowledge and domain knowledge, tutoring strategies, and communication [3]. Nevertheless, the perception-planning-action model is more favourable for building a robot system from the design point of view [17].

As is illustrated in Figure 1, the perception module adopts multi-modal sensors to observe students' activities, and uses AI techniques for learning style and knowledge mastery analysis, serving the input for the planning module. The planning module builds internal models for students, and evaluates teaching outcomes for different teaching strategies before making a decision. According to the teaching decision, the action module constructs teaching-learning scenes to generate appropriate social and physical milieus, and uses multi-modal communication channels to deliver teaching contents. Continuous feedback enables the perception-planning-action loop to perform online adaptation and endows the ITR to learn [17]. Even so, due to the complexity and ethical issues of teaching pedagogy and knowledge structures of the curriculum, human tutors may monitor and intervene in the perception-planning-action loop during either design or runtime process.

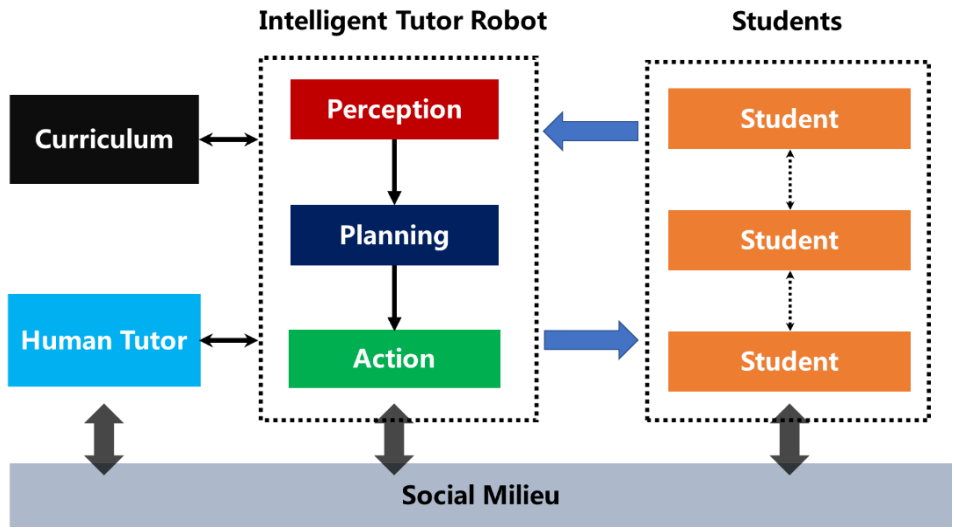

Figure 1 Perception-Planning-Action Architecture of the Intelligent Tutor Robot (ITR) and its Interactions with Human-Tutor, Social Milieu, Curriculum and Students.

*3.1. Multi-Modal Perception of Students*

For the human-tutor, the foundation of effective teaching that leads to student learning is to recognise an individual student's learning style and level of knowledge [18]. Firstly, the ITR may be equipped with audio-visual, tactile, inertial sensors to capture multi-modal data from external environments, from which the students' activities may be analysed. Also, with the aid of e-learning techniques, the surveys and questionnaires may be used for students' self-evaluation in the modal form of text. The multi-modal data should be fused before extracting useful information about the students learning status, learning styles and knowledge level, where AI techniques may be applied [6].

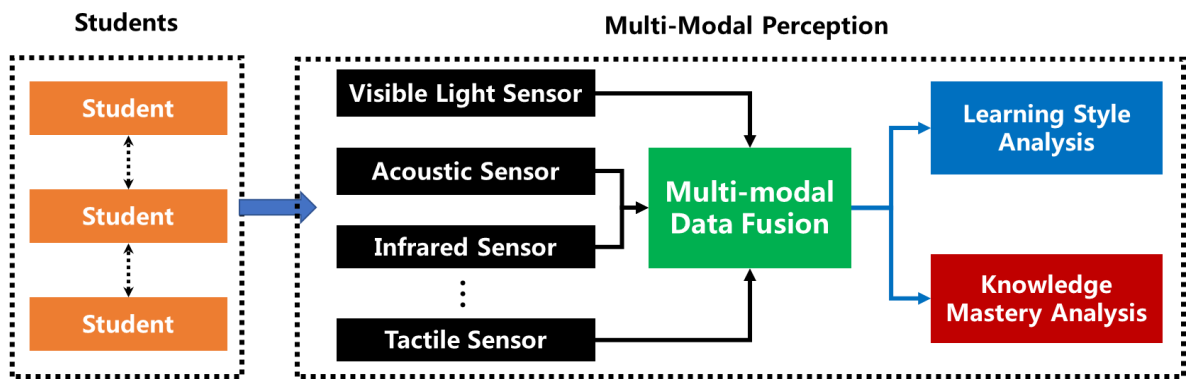

Figure 2 Architecture of Multi-modal Perception with Data Fusion and Analysis.

Multi-modal Data Fusion

The first phase of initiating an effective cognitive perception-planning-action loop is to acquire sufficient information concerning the environments and the students. As shown in Figure 2, the ITR may use multi-modal perception by utilising multiple information channels, e.g. audio, visual, tactile, electromagnetic, electroencephalograph sensors, etc. These sensors may outperform and augment human-tutor sensory capabilities. Pre-processing steps are demanded before extracting information from the raw signals. Noise and interference are to be removed while preserving the useful information, then signals and data gleaned from multiple independent channels should be aligned in the spatial-temporal space to avoid incorrect correlations [19].

In the context of the ITR design, multi-modal data fusion is not specifically explored, but the topic is well-investigated and still a hot one in the context of data-mining and robotics. Due to space limitations, the rest of the section would not aim at a thorough review of the state-of-art in data fusion.

Instead, it reviews three recent application driven surveys on data fusion [20, 21, 22] and their great potential applications in intelligent tutor design.

- *Pixel-level data fusion*: In [20], the authors provide a review of contributions on pixel-level data fusion for multiple visual sensors, with applications to remote sensing, medical diagnosis, surveillance and photography. Although pixel-level data fusion is not devoted to intelligent educational applications, the fusion methods and quality measures design may be transferred to the design of ITRs. For example, matting methods may be used for multiple image fusion of moving objects captured by the moving visual sensor mounted on the ITR in dynamic scenarios.
- *Human action recognition*: In [21], a survey focusing on data fusion in applications of human action recognition is given. The authors compare the pros and cons of adding inertial sensory data to traditional visual sensors. Human action recognition may find its wide range of applications such as video analytics, robotics and human-computer interaction, therefore it may be directly adopted in intelligent tutor design. For instance, in a complex tutoring scenario of much occlusion and moving objects, the inertial sensory data may complement the visual sensory data with a limited field of view using support a vector machine and hidden-Markov model.
- *Affective computing*: In [22], the authors discuss in detail the recent contributions in affective computing, and the methods are evolving from uni-modal analysis to multi-modal fusion. Affective computing is an inter-disciplinary research field devoting to endow machines with cognitive capabilities to recognise, interpret and express emotions and sentiments, hence it may be applied directly in ITR for students' emotion and sentiment interpretation that helps the analysis of learning style and knowledge level. From audio-visual sensory data, the ITR may facilitate a series of metabolic variables, e.g. heart monitoring and eye tracking, to gather information about the emotion as well as the level of engagement and attention of the students. Multiple kernel learning and deep convolutional neural network methods may be adopted for the sentiment detection of students.

An important issue should be noted that the increasing amount of perceptual data may overwhelm the following stages of information processing, therefore it demands teaching task-oriented compression. However, as found by the recent survey [6], there are several predictors that have been taken into account different sets of variables, none of the papers manages to compare the effectiveness of different attributes in predicting learning styles. Hence, it remains an open question on how to select the appropriate set of variables. A promising way of salient feature design is the attention mechanism, which has been shown effective in handling a large amount of audio-visual data flows, striking a beneficial balance between information preservation and computing efficiency [23, 24].

Learning Style and Knowledge Mastery Analysis

With the aid of multi-modal data fusion, the ITR may gather data about the students with external-assessment and self-assessment, and then implement statistical machine learning methods to analyse the students' learning style and knowledge level.

Before delving into the external and self-assessment methods, the ITR selects the criteria for learning style. Although a wide range of learning style theories have been developed in the literature, where most of them cannot be compared quantitatively so that it is hard to argue one of them could outperform others [6]. Therefore, the human-tutor may have to analyse the curriculum and select the learning style theory for the ITR and most importantly, the learning style theory should be projected to a well-defined model in terms of categorisation and quantification. For example, Felder–Silverman's model [25] proposes the well-known four-dimension model for learning styles: perception (Sensory/Intuitive), information input (Image or Verbal), information process (Active or Reflective) and understanding (Sequential or Global); each dimension may be labelled in supervised learning and be quantitatively scaled.

After selecting the learning style model, external-assessment may be implemented during the teaching process, which is motivated by human-tutors' capabilities of observing students' activities and affections to assess learning status. By analysing the heterogeneous data gathered from multi-modal sensors, machine learning techniques may extract hidden modes of students' learning style and status affected by personality, motivation and emotion [22]. Students' self-assessment that incorporates levels of competences for each concept or skill may also be implemented, which allows the ITR to continuously calibrate planning and actions to better meet student needs. In this case, a simple way for self-assessment questionnaires is to use a scale that ranges from cursory level, factual knowledge level, to conceptual knowledge level and application level. With the aid of simple statistical machine learning techniques and visualised analytics like bar charts, radar charts and ranking tables, information about per-student knowledge mastery level may be extracted [26, 27, 28].

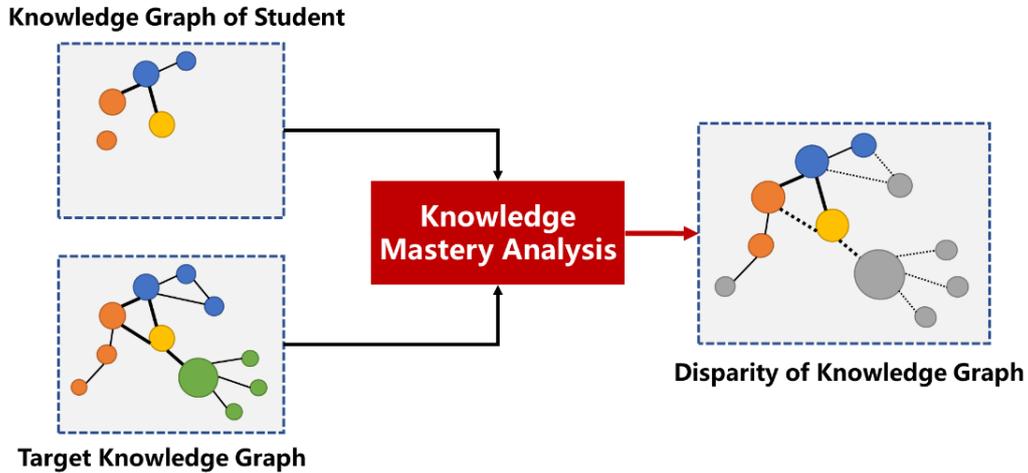

Figure 3 Disparity analysis between the knowledge graph of a student and that of an expert.

More sophisticated concept maps may be adopted to analyse students' level and organisation of knowledge. The knowledge along with its structure for mastering a curriculum heavily relies on the evaluation and experience of human-tutors. Hence, the ITR may receive input from human-tutors, and a well-designed and formalised data structure for the human-robot tutor interface is the concept map and knowledge graph [29]. In a concept map or knowledge graph, each vertex is a concept or ontology and a network is formed to describe the relationship between ontologies. Based on a series of related knowledge concepts and a standard concept map designed by human experts, the ITR may request students to construct a knowledge graph [2918] representing everything that they know about the topic, and then use machine learning techniques designed for graph networks [30, 31] to evaluate the density and intensity of the students' knowledge structures. This reveals students' blind spot and weak knowledge connections required to be strengthened during the following-up teaching and learning process. Take an example shown in Figure 3, the knowledge graph may be compared with a target graph for the curriculum by the knowledge level analysis module, so that the disparity graph may be generated for revealing the weak and missing knowledge of a student.

The extracted information indicating students' learning style and knowledge level, may be jointly fed to the planning modules of the ITR.

*3.2. Planning of Teaching Contents and Strategies*

According to the learning style and knowledge level analysis of the students, the ITR may build the student model for each student or a group of students, based on which it may predict short-term and long-term outcomes of delivering certain teaching contents via specific teaching strategies. Then, these predictions may be fed to the decision-making process.

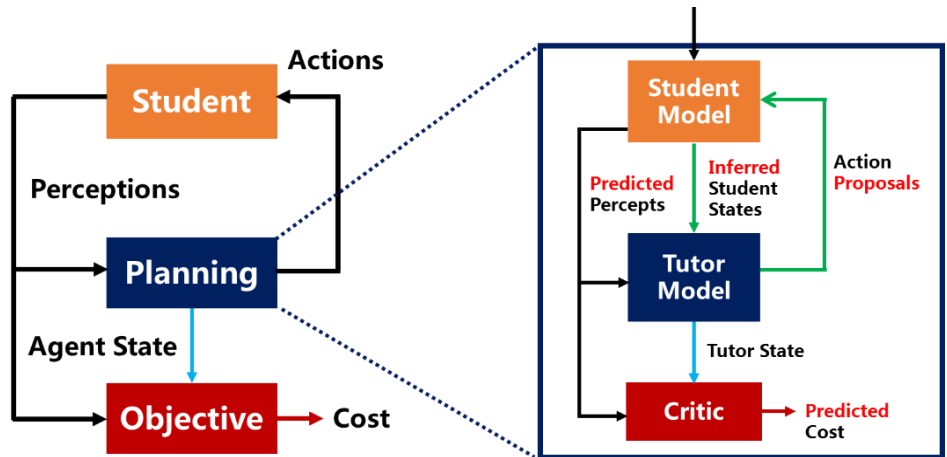

Figure 4 Illustrations of Internal Models of the Planning Module.

Student Model and Teaching Outcome Prediction

The ITR reasons students cognitive and emotional information, then builds the information into the student model. This model includes the dimension of topics, misconceptions, affective characteristic, student experiences and stereotypes, and may be instantiated into operational student modules [3]. Hence, the ITR may evaluate the outcome of potential teaching actions in its own "brain" as seen in Figure 4, before performing actions in the real world [17, 32].

The analytical results from the perception module form the data basis of a student model, but the method of using these data in building a student model may be diversified. Firstly, the model tracing method assumes that students may be modelled as rule-based agents, and the execution trace of these rules is available for the ITR to infer student states [33]. Secondly, the constraint-based model method assumes that learning cannot be fully recorded and only errors can be recognised by an ITR, which may build an annotated domain model indicating the gap between students' knowledge and expert's knowledge in the context of the curriculum, as well as a bug library indicating the misconceptions and missing knowledge of the students [34]. Thirdly, machine learning method avoids necessities on the full model of student behaviours and such techniques used for modeling student knowledge most often include statistical inference, and have been used in the ITR to predict how and when student responses and whether the response is expected to be correct [3, 35]. However, machine learning methods lacks causality analysis and interpretability [16, 33, 35].

The distinctions between the model-tracing, constraint-based and machine learning methods are vanishing. For example, a classical model-tracing based ITR called PAT represents human declarative knowledge by modular units called chunks and human procedural knowledge by if-then production rules for capturing the students' weakness and misconceptions [3]. In comparison to hand-crafted modelling in PAT, a more recent model-tracing based ITR in [36] adopts deep recursive neural network, a machine learning model to avoid explicit encoding of human domain knowledge, and can capture complex representations of student knowledge.

Based on the built student model, the ITR may simulate, test and predict teaching outcomes by interacting with student models. As shown in Figure 4, the student model may be instantiated as a virtual agent that reacts to ITR's actions, which allows the ITR to perform and compare different teaching actions. Based on the interactions between the student models, the ITR may adjust its strategies to the students' particular feedback.

Teaching Decision Making

The ITR may interact with the student's model with multiple candidate plans of teaching contents and strategies, and make predictions on the outcome of each candidate plan. The next step is to compare these candidate plans and make a decision.

In general, there is no optimal plan as the solution may serve contradictory objectives in multiple dimensions. For example, the plan may select easier problems for motivating students to enjoy short-term success, while it may lead to smaller learning gains in the long run. As shown in Figure 5, the ITR has to take at least three factors into account, e.g. long-term education objective, short-term curriculum objective and student personal objective. The long-term education objectives include facilitating students into self-directed learners, equipping students with general learning tools so that the students excel in the life-long learning progress. The short-term curriculum objectives usually aim at promote student to master knowledge in the context of the curriculum. The student personal objective may be further diversified according to their own interests [18].

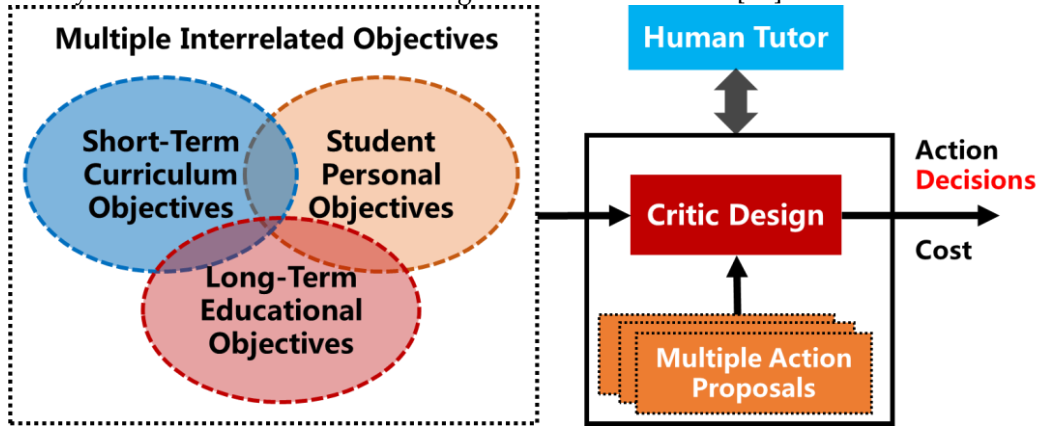

Figure 5 Critic design needs to take into account of multiple interrelated objectives and may benefit from human tutor supervision. The critic may select from multiple action proposals before making a decision.

In many cases, it is a challenge for the ITR to formulate the multiple-objective optimisation problems by itself. As shown in Figure 5, human tutors and curriculum designers may intervene and design quantitative and assessable cost functions to facilitate the decision-making process for the ITR. As long as the cost function is properly designed for the critic, the machine learning techniques may compute the decision actions from multiple action proposals. In [37], the authors propose a selection algorithm for adaptive tests based on multi-criteria decision models integrating expert knowledge by fuzzy linguistic information. In [38], the authors propose a four-stage decision-making plan and a synthesis mechanism for cognitive maps of knowledge diagnostic.

In general, the decision plan may be interpreted in a form of action sequence to intervene in and impact students' learning [3]. The action sequence is fed to the action modules of the ITR. The actions have multiple forms in order to adapt to a variety of teaching-learning situations. Specifically, presentation of examples reduces cognitive load for students in complex problem-solving tasks, and timely feedback provides information for correcting students' errors and misconceptions, meanwhile enhances motivations for higher levels of efforts [18]. The specific design of action sequence relies heavily on selecting tutoring strategies. The strategy may be inferred from either mimicking human teaching such as apprenticeship [39], or be based on a variety of learning theories such as, cognitive learning [40], constructivism [41] and situated learning [42].

*3.3. Action through Multi-modal Communications*

After receiving the plan result, the action module resolves the action sequence indicating what, when and how teaching contents are to be delivered to the students. Even with the best students and teaching plan, an ITR is of limited value without effective communicative strategies. Therefore, the action module needs to form an appropriate scenario for the students, as well as to provide effective communication channels for transmitting the teaching contents and receiving feedback from students. Various techniques and demos have been proposed to enable scene construction and multi-modal communication channels in the action module.

Scene Construction

Physical or virtual scenes generated by the ITR may allow students to immerse the learning process and use multi-modal perception to acquire information, which may lead to more effective learning. As shown in Figure 6, the virtual scene generation is of much higher flexibility and relatively low costs, therefore becomes a major research direction.

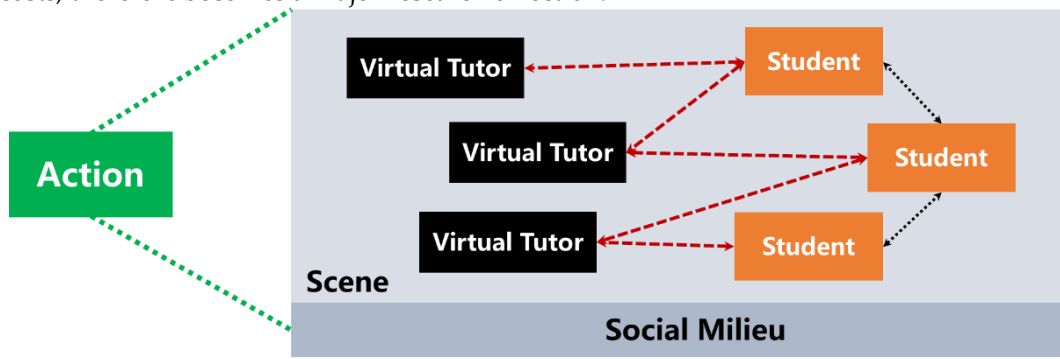

Figure 6 Action module constructs an immersive scene for students with multiple virtual tutors.

The action module may generate virtual-reality and augmented-reality environments, including the social milieu and physical milieu powered by a physical engine. In [43] and [44], the authors provide a thorough review of using virtual reality techniques to build virtual environments for laboratory and training facilities. Recent advances in applying augmented reality techniques in education have been reviewed in [45-47]. Furthermore, a comparison of the pros and cons in using virtual-reality and augmented-reality for learning system design is given by [48].

Then, the action module may inhabit virtual tutors in the virtual-reality environment. The action module may imitate human tutors who have natural behaviours such as natural languages, gestures, facial expressions to interact with students [49-51]. Furthermore, the action module may enable collaboration and communication in ways that are impossible with traditional tutoring because of ethical issues, e.g. virtual patients [52].

Multi-modal Communication Channel

Within the physical or virtual scenes generated by the ITRs, multi-modal forward channels are built for delivering the action sequence. It has been found that rather than passively receive knowledge from the tutor, the students construct their own structures and organise their own knowledge. Therefore, the ITR should promote critical thinking, self-directed learning, and self-explanation via social communication exploiting both student affection and facial features.

By simulating human communicative strategies, graphic communication and natural language are the conventional communication channels for delivering social communication. Although a vast number of contributions have been devoted [3, 16], it is worth noting that recent work on generated adversarial networks allow high-fidelity generation of facial expression [53, 54] and natural language [55, 56].

The physically embodied robots may endow physical and emotional interactions, increasing cognitive and affective outcomes, and have achieved outcomes similar to those of human tutoring on restricted tasks [57]. The authors of [58, 59] assess the effect of the physical presence of a ITR in an automated tutoring interaction, showing that physical embodiment and personalisation can yield significant benefits in educational human-robot interactions, hence producing measurable learning gains. The authors of [60, 61] explore the social robot capability of improving students' curiosity, while those of [62] use hints and distractions of curious facts in improving students learning performance to explore social robots.

**4. Discussion and Conclusions**

This paper transforms the relationship model describing the teaching-learning process into the well-established perception-planning-action ITR model in the area of AI.

- The first phase of initiating an effective perception-planning-action loop of ITR is to acquire sufficient information concerning the environments and the students, so it introduces multi-modal perception that utilises multiple communication channels. Then, the ITR leads the students to do a variety of activities for external-assessment and self-assessment, gather information, and then implement statistical machine learning methods to analyse students' learning style and knowledge level.
- According to students' learning style and knowledge level, the second phase of ITR builds the student model for each student or a group of students, based on which it can predict short-term and long-term outcomes of delivering certain teaching contents via specific teaching strategies. Then, the ITR may interact with the student's model with multiple candidate plans of teaching contents and strategies, and make predictions on the outcome of each candidate plan. Human tutors and curriculum designers may also intervene in the decision making.
- After receiving the plan result, the third phase is activated by the action module of the ITR that resolves the action sequence indicating what, when and how teaching contents are to be delivered to the students. The action module needs to form an appropriate scenario for the students, as well as to provide effective communication channels for transmitting the teaching contents and receiving the feedback from students, using scene construction and multi-modal communication channels in the action module.

With the rapid progress of AI techniques, many open research areas may be defined for ITRs with the aid of the perception-planning-action model.
- Perception: Multi-modal data fusion is not fully researched in the context of ITR design, even though the topic is well-investigated and still in heated discussion in the field of data-mining and robotics. Also, little work has compared the effects of different factors that indicate learning styles, hence it remains an open question on how to select the appropriate set of variables to predict students learning process.
- Planning: In the context of student modelling, the model tracing, constrained model and machine learning methods have their own application scenarios and limitations, therefore the recent advances of explainable AI may be incorporated in ITRs. In terms of decision making, limited research have contributed to the problem formulation that take into account the multi-objective problem, namely, jointly considering the long-term education objective, short-term curriculum objective and student personal objective.
- Action: The action module field turns out to be the most researched area with the state-of-art AI techniques. Even so, the applications of advanced AI techniques in ITRs such as generated adversarial networks are underway, while the design of physically embodied robots is far from mature.